\documentclass[letterpaper, 10 pt, conference]{ieeeconf}  

\IEEEoverridecommandlockouts                              

\usepackage{graphicx}
\usepackage{placeins}
\usepackage{capt-of}
\usepackage{amsmath} 
\usepackage{amssymb}  
\usepackage{booktabs}
\usepackage{makecell}

\title{\LARGE \bf
DEFT-VLA: Stage-Localized Force Specialization for Contact-Rich Robotic Manipulation
}

\author{
Zan Huang$^{1,\dagger}$,
Yibo Qiu$^{1,\dagger}$,
Kan Ding$^{1,\dagger}$,
Xuanping Zhang$^{2}$,
Ao Tian$^{1}$,
Ronald Xu$^{1}$,
and Mingzhai Sun$^{1,*}$%
\thanks{This work was supported by the Gusu Leading Talent Entrepreneurship
and Innovation Program (Grant No. ZL2024349).}%
\thanks{$^{\dagger}$Zan Huang, Yibo Qiu, and Kan Ding contributed equally
to this work.}%
\thanks{$^{1}$Zan Huang, Yibo Qiu, Kan Ding, Ao Tian, Ronald Xu,
and Mingzhai Sun are with the Suzhou Institute for Advanced Research,
University of Science and Technology of China, Suzhou, Jiangsu, China,
and also with the School of Biomedical Engineering, Division of Life Sciences
and Medicine, University of Science and Technology of China,
Hefei, Anhui, China.}%
\thanks{$^{2}$Xuanping Zhang is with the School of Robotics,
Xi'an Jiaotong-Liverpool University, Suzhou, Jiangsu, China.}%
\thanks{$^{*}$Corresponding author: Mingzhai Sun
(e-mail: {\tt\small mingzhai@ustc.edu.cn}).}%
}

\IEEEaftertitletext{%
\vspace{-0.5em}
\begin{minipage}{\textwidth}
    \centering
    \includegraphics[width=0.9\textwidth]
    {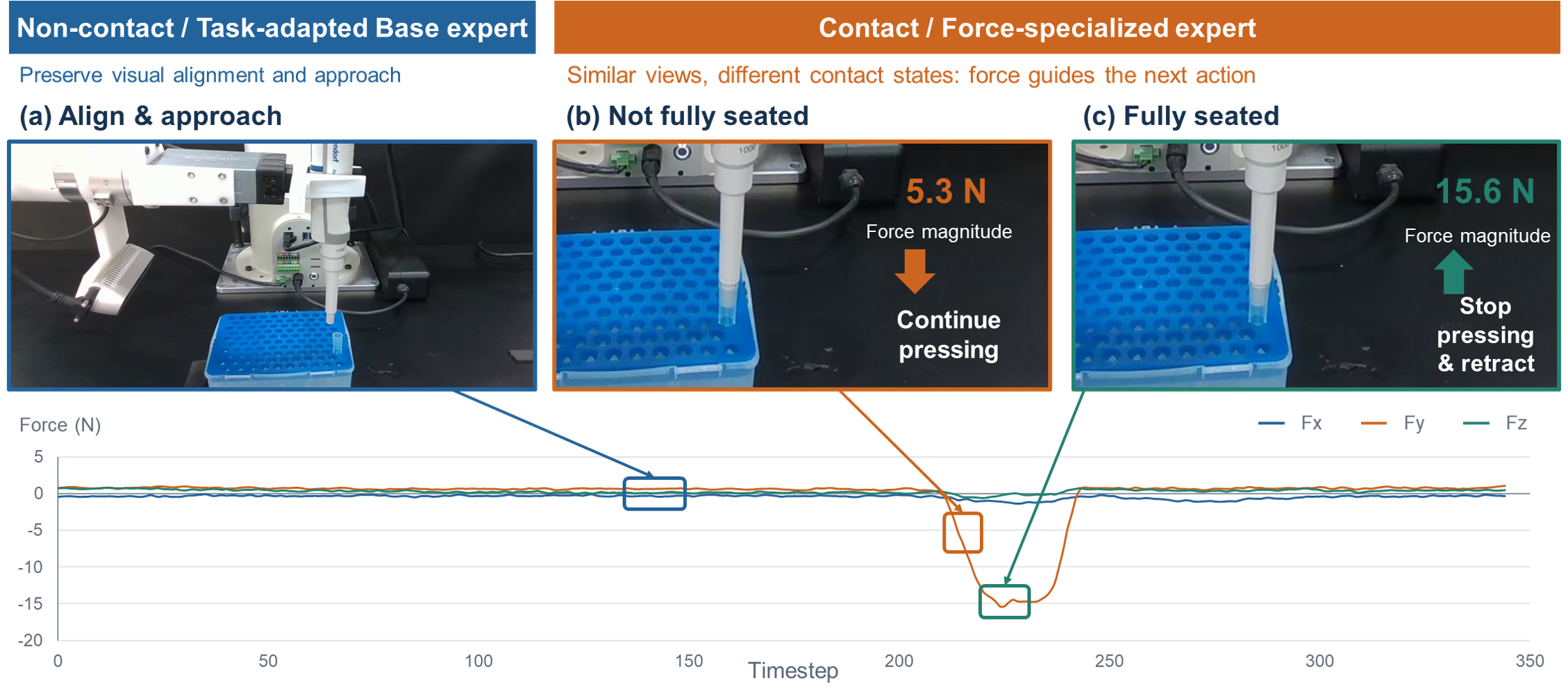}
    \captionof{figure}{
    Overview of DEFT-VLA. The task-adapted base expert preserves
    vision-dominated alignment and approach during non-contact
    stages (a). During contact, visually similar observations may
    correspond to different physical states. Guided by force
    feedback, the force-specialized contact expert continues
    pressing before the pipette tip is fully seated (b), but stops
    and retracts after full seating (c). The force traces illustrate
    the corresponding stage-dependent interaction signals.
    }
    \label{fig:overview}
\end{minipage}
\vspace{0.5em}
}

\begin{document}
\bstctlcite{IEEEexample:BSTcontrol}

\maketitle
\thispagestyle{empty}
\pagestyle{empty}

\begin{abstract}

Contact-rich manipulation poses a stage-dependent sensory conflict: vision and proprioception are essential for localization and approach, when force is largely uninformative and potentially distracting, whereas force becomes critical after contact, when irrelevant visual details and correlated robot states may become shortcuts. We propose \textbf{DEFT-VLA} (\textbf{D}ual-\textbf{E}xpert \textbf{F}orce-specialized \textbf{T}raining for Vision-Language-Action Manipulation), which retains the task-adapted base expert for vision-dominated non-contact behavior and introduces a force-aware contact expert for critical contact stages, localizing force-conditioned adaptation to where it is needed. We further introduce a \emph{contact-localized force-attending curriculum} that applies force-attending post-training only to the contact expert, progressively restoring vision and masking state tokens to suppress shortcuts. Real-robot experiments on pipette tip attachment, bimanual pipette aspiration, and whiteboard wiping show that DEFT-VLA effectively exploits force feedback to improve contact-stage performance while maintaining strong non-contact behavior and robustness to held-out grasp heights. It improves overall task success by 20--40 percentage points over the strongest baseline on each task.

\end{abstract}


\section{INTRODUCTION}

Contact-rich manipulation requires different sensory information at different stages. During non-contact behaviors such as localization, alignment, and approach, vision and proprioception provide the primary spatial cues. In contrast, before physical interaction, wrist force/torque measurements contain little task-relevant contact information and may be dominated by sensor noise or incidental dynamics, making indiscriminate force conditioning potentially disruptive to visuomotor behavior~\cite{he2025foar}. Recent vision-language-action (VLA) models provide strong semantic and visuomotor priors for these vision-dominated stages~\cite{brohan2023rt2,kim2025openvla,black2024pi0,black2025pi05}. Once contact begins, however, sensory relevance shifts: visually similar configurations can correspond to different physical states, such as insufficient contact, jamming, or successful completion. Force/torque feedback helps distinguish these states and guide corrective actions when visual observations alone are insufficient~\cite{liu2025forcemimic,he2025foar,xue2025reactive}. Fig.~\ref{fig:overview} illustrates this distinction: vision guides pipette alignment and approach, while force feedback determines whether to continue pressing or retract after the tip is seated.

Existing force-aware approaches address this sensory transition from two complementary perspectives. One line of work investigates \emph{when} force should influence the policy. FoAR modulates force fusion using predicted future contact, while Reactive Diffusion Policy introduces a fast tactile- or force-conditioned loop for responsive contact control~\cite{he2025foar,xue2025reactive}. Another line studies \emph{how} to make force behaviorally useful. ForceVLA uses phase-aware FVLMoE fusion during action generation, whereas FACTR and CRAFT explicitly address the tendency of policies to neglect physical feedback when alternative sensory shortcuts are available~\cite{yu2025forcevla,liu2025factr,zhang2026craft}. During contact, visually salient but interaction-irrelevant details and robot states correlated with demonstration progress can still dominate action prediction. These two challenges, however, are largely treated separately: phase-aware fusion controls when force enters the policy, whereas force-attending methods alter modality reliance without localizing force-specialized optimization to a dedicated expert or contact-stage data. This raises two questions: can force specialization be confined to contact-critical stages while preserving the task-adapted policy for vision-dominated behavior? Within such a specialist, can both visual and proprioceptive shortcuts be suppressed so that force becomes behaviorally useful rather than merely available?

We introduce \textbf{DEFT-VLA}
(\textbf{D}ual-\textbf{E}xpert \textbf{F}orce-specialized
\textbf{T}raining for Vision-Language-Action Manipulation),
a stage-specialized framework for force-aware VLA adaptation.
We first adapt a pretrained VLA policy to the target task using complete demonstrations and retain the resulting task-adapted base expert for vision-dominated non-contact behavior. A force-aware contact expert is initialized from the base expert and subsequently specialized using only contact-stage data, while the shared vision-language backbone and base expert remain frozen. During execution, a lightweight stage-aware router selects the active expert. We further propose a \emph{contact-localized force-attending curriculum}. Building on FACTR's visual-blurring curriculum~\cite{liu2025factr}, it confines force-oriented post-training to the contact expert and introduces state-token masking to reduce proprioceptive shortcuts correlated with task progress. Visual suppression and state masking gradually decay during training, while language and force inputs remain available. The dual-expert architecture determines where force specialization occurs, whereas the curriculum encourages the contact expert to exploit force once physical feedback becomes relevant.

We evaluate DEFT-VLA on three real-robot tasks: pipette tip attachment, bimanual pipette aspiration, and whiteboard wiping. Our experiments measure overall and stage-wise success, evaluate generalization to held-out pipette grasp heights, and examine the contribution of state-token masking. DEFT-VLA improves overall task success by 20--40 percentage points over the strongest evaluated baseline on each task, with gains concentrated in the contact stage while maintaining strong non-contact performance. Our contributions are threefold:
\begin{itemize}
    \item We introduce a stage-specialized dual-expert VLA framework that retains a task-adapted base expert for non-contact behavior and trains a separate force-aware expert on contact-stage demonstrations.
    \item We propose a contact-localized force-attending curriculum with visual suppression and state-token masking to reduce visual and proprioceptive shortcuts during contact.
    \item We demonstrate improved task success across three real-robot
    contact-rich tasks, together with generalization to held-out pipette
    grasp heights and a state-token masking ablation.
\end{itemize}

\begin{figure*}[t]
    \centering
    \includegraphics[width=\textwidth]{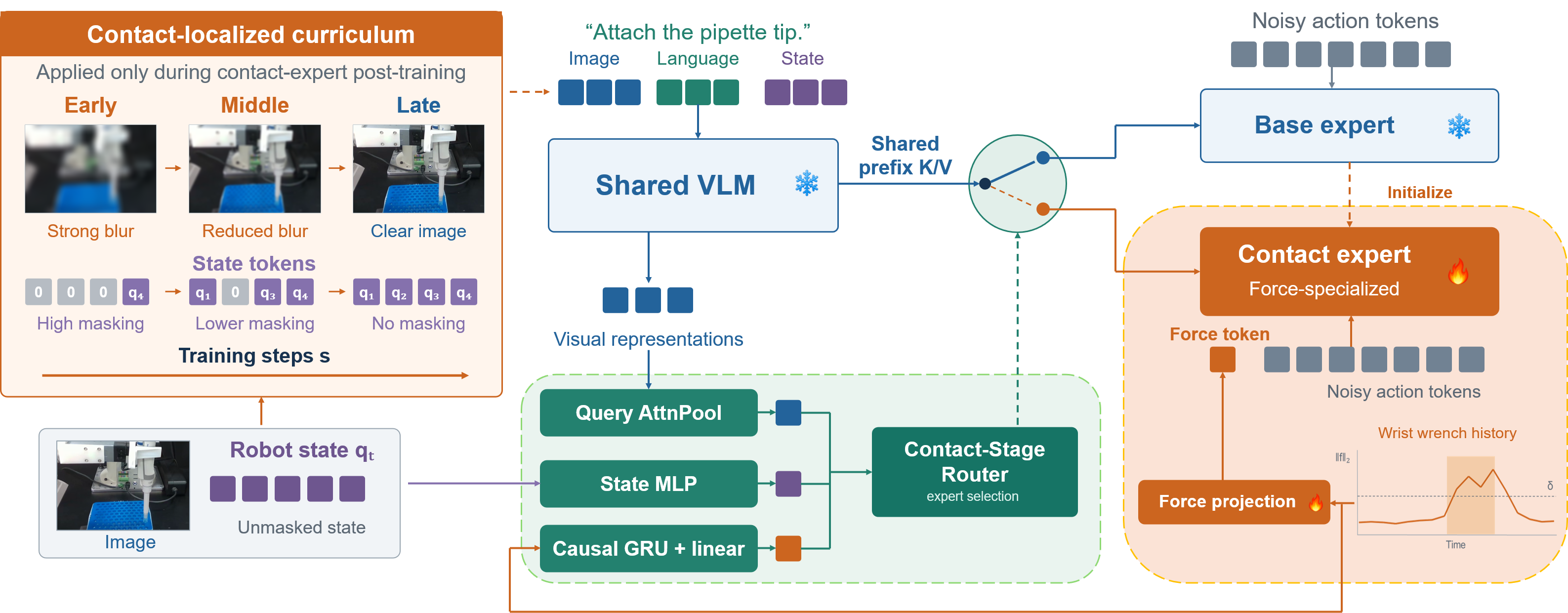}
    \caption{
    Overview of DEFT-VLA.
    A shared VLM conditions a task-adapted base expert for non-contact behavior and a force-specialized contact expert for critical contact stages.
    The contact expert is initialized from the base expert and receives an additional force token.
    During contact-expert post-training, the contact-localized curriculum gradually restores image clarity and the original state tokens to suppress visual and proprioceptive shortcuts.
    At inference, a lightweight stage estimator fuses task-conditioned visual features, robot state, and causal wrench history, while a contact-stage router selects the active expert.
    Snowflake and flame symbols denote frozen and trainable modules, respectively.
    }
    \label{fig:method}
\end{figure*}

\section{Related Work}

\subsection{Generalist Vision-Language-Action Models}

Generalist VLA models build on scalable and cross-embodiment robot learning. RT-1 demonstrated large-scale multi-task visuomotor control, while Open X-Embodiment, RT-X, and Octo extended policy learning across heterogeneous embodiments~\cite{brohan2023rt1,openx2024rtx,ghosh2024octo}. Building on these advances, RT-2, OpenVLA, and the $\pi_0$ family leverage visual-language priors to learn transferable visuomotor policies~\cite{brohan2023rt2,kim2025openvla,black2024pi0,black2025pi05}. These priors support vision-dominated behaviors such as localization and approach, but contact-rich execution may depend on physical cues that are weakly observable from vision. This motivates force-aware adaptation that preserves pretrained visuomotor competence.

\subsection{Force-Aware Learning for Contact-Rich Manipulation}

Action-chunking and diffusion-based policies provide strong backbones for temporally coherent imitation learning~\cite{zhao2023act,chi2023diffusion}. ForceMimic, FoAR, Reactive Diffusion Policy, and Adaptive Compliance Policy respectively study force-centric imitation, phase-aware fusion, high-frequency physical feedback, and compliance learning~\cite{liu2025forcemimic,he2025foar,xue2025reactive,hou2024adaptivecompliance}. Related tactile and visuotactile policies improve direct contact sensing~\cite{tacdiffusion,visualforcetactile,multimodalforcematched}, while Tactile-VLA and TacVLA incorporate tactile feedback into language-conditioned action generation~\cite{huang2025tactilevla,zhang2026tacvla}. Recent force-aware VLA extensions use torque tokens, hybrid force-position control, or fast-slow force adaptation~\cite{zhang2025tavla,li2026forcevla2,li2026favla}. ForceVLA integrates projected real-time 6D force into its action expert and uses FVLMoE for modality- and phase-aware routing within a single policy~\cite{yu2025forcevla}. Unlike this intra-policy routing, DEFT-VLA explicitly switches between separately parameterized base and contact action experts.

Providing physical feedback does not ensure its use, because correlated observations can become shortcuts in imitation learning~\cite{dehaan2019causalconfusion}. FACTR and CRAFT address this problem through force-attending curricula, while FACTR~2 further studies force-aware learning across pre-contact and contact segments~\cite{liu2025factr,zhang2026craft,oh2026factr2}. Architecturally, Force Policy separates vision-guided free-space motion from high-frequency contact control, whereas LIFT adds a force-conditioned reactive expert initialized from the original action expert~\cite{fang2026forcepolicy,wang2026lift}. PriorVLA preserves a frozen prior expert, while Knowledge Insulation protects pretrained VLM knowledge during action-expert training~\cite{guo2026priorvla,driess2025knowledgeinsulation}. FM-VLA and FWBC-VLA instead inject force histories or contact estimates into action experts~\cite{li2026fmvla,zhang2026fwbcvla}. Complementary approaches adapt contact behavior through corrective intervention or tactile-guided online reinforcement learning~\cite{forceaware_dagger,zheng2026torlvla}. DEFT-VLA differs from these approaches by combining force-thresholded
contact-stage expert training, learned hysteretic routing, and a
contact-localized curriculum that suppresses visual and proprioceptive
shortcuts.

\section{METHOD}
\label{sec:method}

Fig.~\ref{fig:method} summarizes DEFT-VLA, including the
stage-localized dual-expert architecture, the contact-localized
force-attending curriculum, and automatic expert routing.

\subsection{Stage-Localized Dual-Expert Architecture}
\label{sec:dual_expert}

Contact-rich manipulation involves stages with distinct sensory
requirements. Localization, alignment, and approach primarily rely on
vision and proprioception, whereas force feedback becomes important
after physical contact for determining whether to continue, stop, or
retract~\cite{he2025foar,fang2026forcepolicy}. Based on this observation, DEFT-VLA retains a task-adapted
base expert for vision-dominated non-contact behavior and introduces a
force-specialized contact expert for critical contact stages.

Building on $\pi_{0.5}$~\cite{black2025pi05}, we construct a policy
with a shared multimodal VLM and two action experts with independent
action-generation parameters. Given image observations $I_t$, language
instruction $l$, and robot state $q_t$, the shared multimodal prefix is
\begin{equation}
    X_t^P
    =
    \left[
    X_t^I,X^l,X_t^q
    \right],
    \label{eq:multimodal_prefix}
\end{equation}
where the three terms denote image, language, and serialized state
tokens, respectively. Both action experts read the shared multimodal
prefix through the original layer-wise attention mechanism, while
only the contact expert additionally receives force information.

The model is trained in two stages. First, the original
$\pi_{0.5}$ policy is adapted to the target task using complete task
demonstrations, producing a task-adapted VLM and base expert. Second,
both components are frozen, and a new contact expert is initialized
from the task-adapted base expert. Only the contact expert and its
force-input module are subsequently updated using contact-stage data.
This design assigns non-contact and force-sensitive contact behavior
to separate experts, preventing force-oriented adaptation from
directly modifying the visuomotor competence acquired by the
task-adapted base expert.

\subsection{Force-Specialized Contact Expert}
\label{sec:contact_expert}

Let $w_t$ denote the measured wrist force/torque observation at
timestep $t$, and let $f_t$ denote its force component. The contact expert receives a sampled recent wrench history
$\mathcal{F}_t^C\in\mathbb{R}^{K_C\times d_w}$, where $K_C$ is the
number of samples and $d_w$ is the wrench dimension.
Similar to token-based physical-feedback conditioning in torque-aware
VLA models~\cite{zhang2025tavla}, the normalized history is flattened
and linearly projected into a force token:
\begin{equation}
    z_t^F
    =
    W_F
    \operatorname{vec}
    \left(
    \operatorname{Norm}
    \left(
    \mathcal{F}_t^C
    \right)
    \right)
    +b_F.
    \label{eq:force_token}
\end{equation}

At flow-matching timestep $\tau$, the contact-expert action suffix is
\begin{equation}
    X_{t,\tau}^C
    =
    \left[
    z_t^F,
    z_{t,\tau}^{a,1},
    \ldots,
    z_{t,\tau}^{a,H}
    \right],
    \label{eq:contact_suffix}
\end{equation}
where $z_{t,\tau}^{a,i}$ denotes the $i$-th noisy action token in an
action chunk of horizon $H$. The contact-action tokens can therefore
attend to both the shared multimodal prefix and the force token,
whereas the base-expert suffix contains no force token.
Within each denoising forward pass, all action tokens attend to
one another, while neither the prefix nor the force token attends
to the action tokens.

To focus contact-expert training on physical interaction, we extract
critical contact stages from complete demonstrations. For timestep
$t$, let
$\mathcal{W}_t=\{t-m,\ldots,t+n\}$ denote the temporal neighborhood
that extends detected contact by $n$ frames before and $m$ frames
afterward. We assign the stage label
\begin{equation}
    c_t
    =
    \mathbf{1}
    \left[
    \max_{j\in\mathcal{W}_t}
    \left\|f_j\right\|_2
    >
    \delta_{\mathrm{task}}
    \right].
    \label{eq:contact_label}
\end{equation}
The contact-stage dataset is then
\begin{equation}
    \mathcal{D}_C
    =
    \left\{
    \left(
    I_t,l,q_t,\mathcal{F}_t^C,a_t
    \right)
    \mid c_t=1
    \right\}.
    \label{eq:contact_dataset}
\end{equation}
The temporal expansion includes observations around contact onset and
release, retaining the transition context required for expert
switching and contact-action generation.

\noindent\textbf{Automatic expert routing.}
At deployment, a lightweight stage estimator selects between the base
and contact experts. It attention-pools the frozen VLM image-token
representations, encodes the current robot state with an MLP, and
processes the recent wrench history using a causal GRU. The resulting
features are concatenated and passed through a fusion MLP with a
sigmoid output. The estimator is trained using the contact labels
$c_t$, while the VLM and both action experts remain frozen. To avoid rapid switching caused by transient probability fluctuations,
the router uses different thresholds for entering and leaving the
contact stage and switches only when the corresponding decision remains
stable for several consecutive control steps. Upon a switch, the
unexecuted actions from the previous chunk are discarded and regenerated
by the selected expert.

\subsection{Contact-Localized Force-Attending Curriculum}
\label{sec:localized_factr}

Even when force is provided to the contact expert, the policy may rely
on visual details or robot states correlated with demonstration
progress to predict actions, thereby underutilizing online physical
feedback. To suppress these shortcuts, we localize the gradually
relaxed visual-blurring curriculum of FACTR~\cite{liu2025factr} to the
contact expert and further introduce state-token masking.

During early contact-expert post-training, Gaussian blur is applied to
the input image:
\begin{equation}
    \widetilde{I}_t
    =
    I_t*G_{\sigma_s},
    \qquad
    \sigma_s
    =
    \sigma_{\max}\lambda_s,
    \label{eq:visual_blur}
\end{equation}
where $G_{\sigma_s}$ is a Gaussian kernel and $s$ denotes the
optimization step. The curriculum coefficient is
\begin{equation}
    \lambda_s
    =
    \begin{cases}
        1,
        & s<S_w,\\[2pt]
        1-\dfrac{s-S_w}{S_d},
        & S_w\leq s<S_w+S_d,\\[5pt]
        0,
        & s\geq S_w+S_d,
    \end{cases}
    \label{eq:curriculum}
\end{equation}
where $S_w$ is the duration of strong visual suppression and $S_d$ is
the subsequent linear-decay duration. The visual input is thus
gradually restored from strongly blurred observations to the original
images.

To further reduce shortcuts from robot states, we apply state-token
masking to the serialized state tokens. Let $x_{t,j}$ denote the
discrete vocabulary ID of the $j$-th serialized state token. We sample
\begin{equation}
m_{t,j} \sim \operatorname{Bernoulli}(p_s),
\qquad
p_s = p_{\max}\lambda_s,
\label{eq:mask_probability}
\end{equation}
and construct each masked state token as
\begin{equation}
\widetilde{x}_{t,j}
=
\begin{cases}
x_0, & m_{t,j}=1,\\
x_{t,j}, & m_{t,j}=0,
\end{cases}
\label{eq:state_masking}
\end{equation}
where $x_0$ denotes the fixed token with vocabulary index $0$. The
replacement is performed on the discrete token IDs before embedding
lookup; thus, masked positions are mapped to the learned embedding of
vocabulary token $0$, rather than to an all-zero embedding. Language
and structural tokens remain unchanged.

During early post-training, visual detail and a subset of state tokens
are suppressed, while language and force history remain fully
available. As $\lambda_s$ decays, both the visual blur and state-token
masking gradually disappear, restoring the complete observations used
at deployment. Only the contact expert and force-token projection are
optimized during this stage. The shared VLM and task-adapted base
expert remain frozen, and training retains the original flow-matching
objective of $\pi_{0.5}$~\cite{black2025pi05}.

\section{Experiments}

We conduct experiments to answer three questions:
\textbf{RQ1:} Can DEFT-VLA improve overall and stage-wise performance in tasks combining vision-dominated preparation and force-sensitive contact interaction?
\textbf{RQ2:} How do the stage-localized dual-expert architecture, the contact-localized force-attending curriculum, and state-token masking contribute to its performance?
\textbf{RQ3:} Can DEFT-VLA remain robust when the initial tool configuration deviates from the training distribution?

\subsection{Setup}

\begin{figure}[t]
    \centering
    \includegraphics[width=0.9\columnwidth]
    {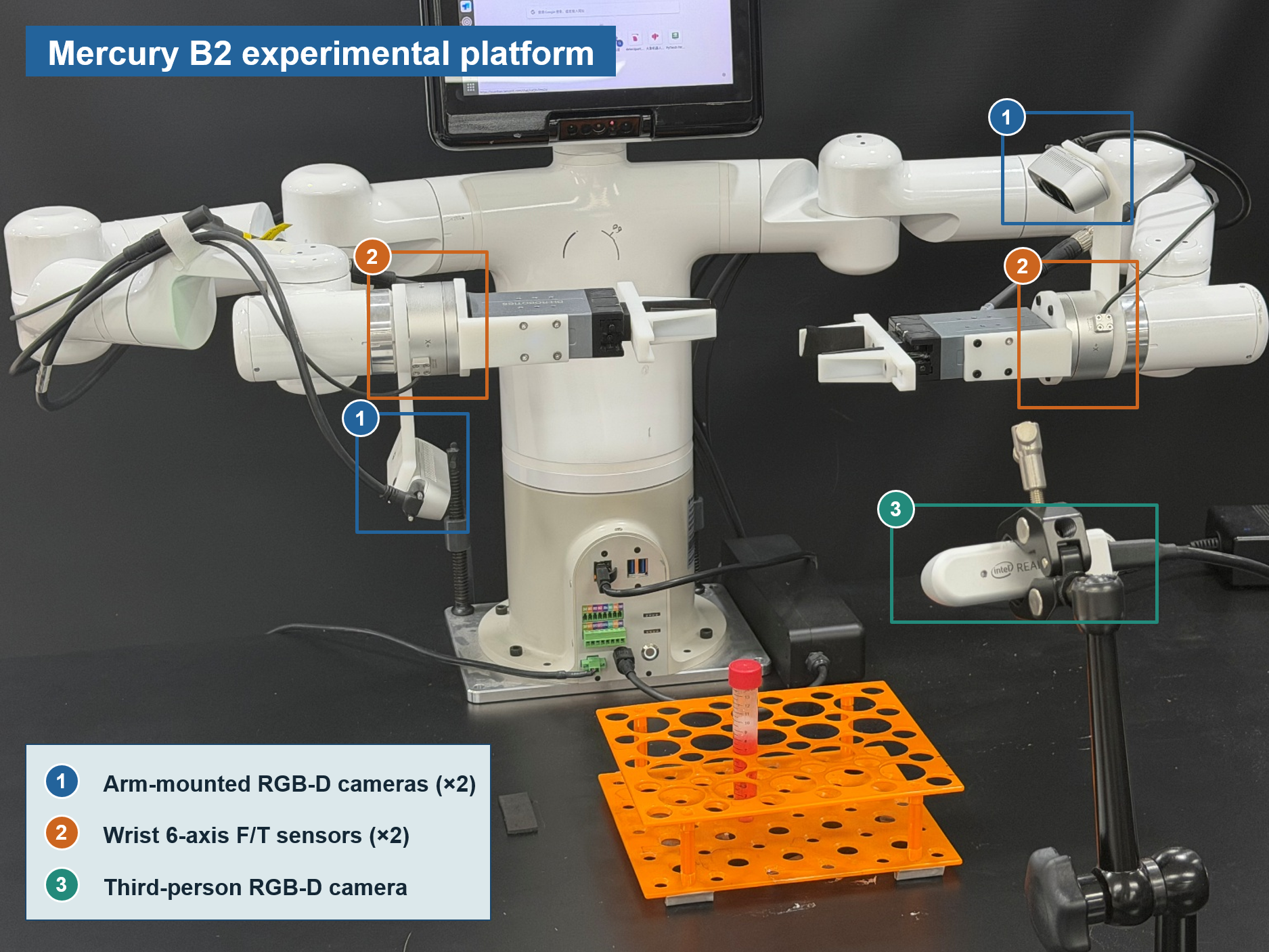}
    \caption{\textbf{Robot platform and sensor configuration.}
    The Mercury B2 dual-arm robot is equipped with two arm-mounted cameras, one third-person camera, and two wrist-mounted six-axis force/torque sensors.}
    \label{fig:platform}
\end{figure}

\noindent\textbf{Robot platform.}
Experiments are conducted on an Elephant Robotics Mercury B2 dual-arm robot operating at 15\,Hz. As shown in Fig.~\ref{fig:platform}, the system is equipped with three Intel RealSense D435 cameras: two arm-mounted cameras providing close-range manipulation views and one external camera providing a third-person view. Each wrist is instrumented with a six-axis force/torque sensor for measuring contact forces and torques.

\begin{figure*}[t]
    \centering
    \includegraphics[width=\textwidth]
    {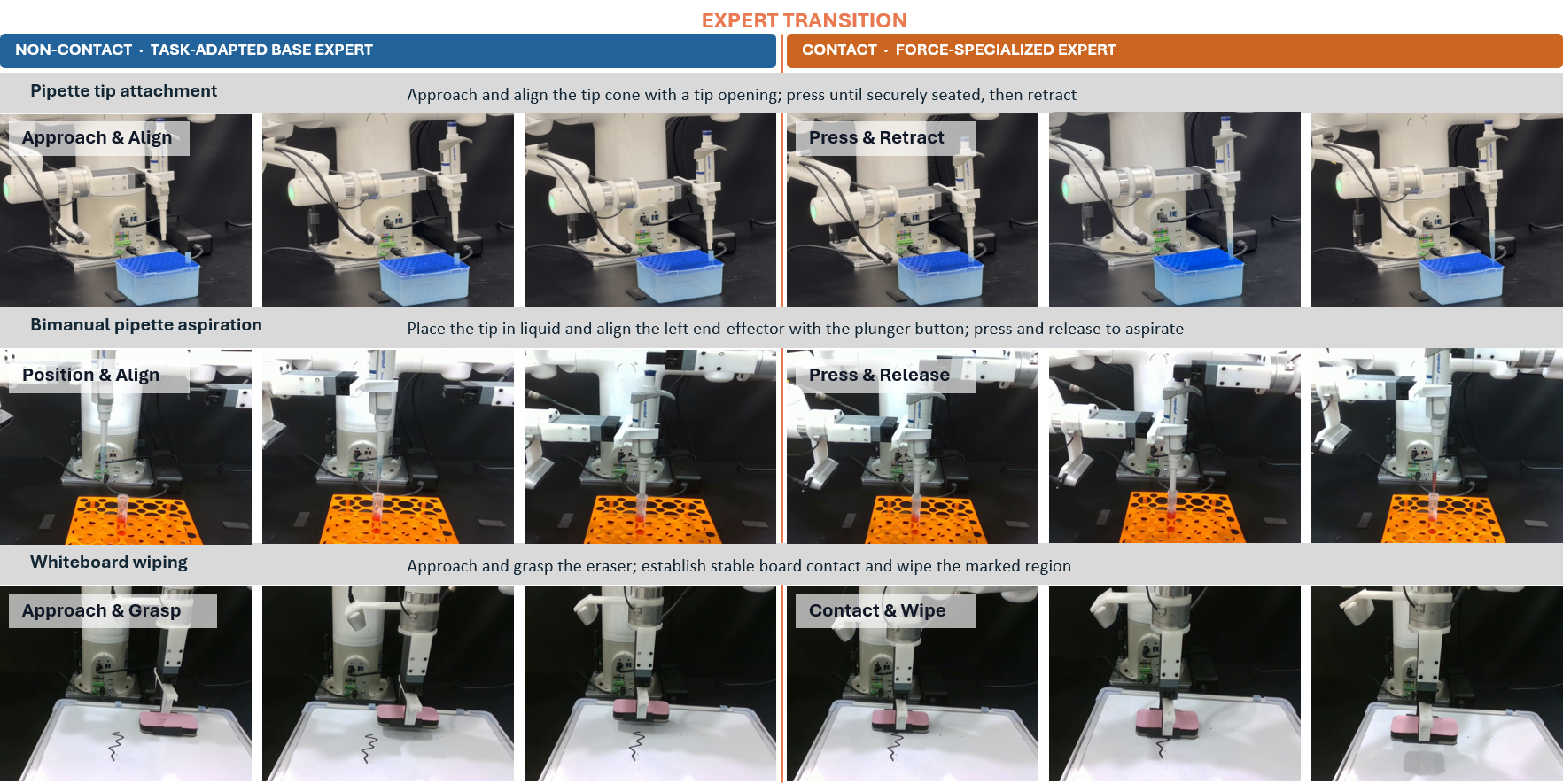}
    \caption{\textbf{Three contact-rich manipulation tasks and their stage decomposition.}
    In each row, the first three frames show the vision-dominated preparatory stage handled by the task-adapted base expert, while the last three show the critical contact stage handled by the force-specialized contact expert. The central boundary denotes the transition from the base expert to the contact expert.}
    \label{fig:tasks}
\end{figure*}

\noindent\textbf{Tasks.}
We evaluate all methods on three contact-rich manipulation tasks. Each task contains a vision-dominated preparatory stage, such as approach, localization, alignment, or tool acquisition, followed by a critical contact stage in which force feedback becomes important. Representative task sequences and their stage decomposition are shown in Fig.~\ref{fig:tasks}.

In \emph{pipette tip attachment}, the robot aligns the pipette tip cone with a target tip, presses until the tip is securely seated, and then retracts. Insufficient force produces a loose attachment, whereas excessive pressing may cause abnormal contact or failure.

In \emph{bimanual pipette aspiration}, the right arm positions the pipette tip in the liquid, while the left arm aligns with, presses, and releases the plunger button to aspirate. Under-pressing may fail to actuate the pipette, whereas over-pressing may trigger the second stop.

In \emph{whiteboard wiping}, the robot grasps an eraser, establishes contact with the marked region, and wipes while maintaining stable normal contact. This task requires sustained force regulation throughout the wiping motion.

\noindent\textbf{Compared methods.}
All methods are built upon task-adapted $\pi_{0.5}$~\cite{black2025pi05}. We compare the following five variants:

\begin{itemize}
    \setlength{\itemsep}{1pt}
    \setlength{\parskip}{0pt}
    \setlength{\parsep}{0pt}

    \item \textbf{$\pi_{0.5}$:}
    the task-adapted policy using image, language, and proprioceptive observations without force input.

    \item \textbf{$\pi_{0.5}$ + Force Token:}
    a single-expert policy that encodes recent wrist force/torque observations as a force token, using a token-injection design similar to TA-VLA~\cite{zhang2025tavla}. 

    \item \textbf{$\pi_{0.5}$ + Force Token + Global FAC:}
    the preceding force-token policy trained with a force-attending curriculum following FACTR~\cite{liu2025factr}. The curriculum is applied globally to the single expert over full-task post-training, providing a direct comparison with our contact-localized adaptation.

    \item \textbf{DEFT-VLA w/o Contact-Localized FAC:}
    the proposed dual-expert architecture with a task-adapted base expert and a force-aware contact expert, where the contact expert is trained without the force-attending curriculum. 

    \item \textbf{DEFT-VLA:}
    our complete method, which restricts force-attending post-training to the contact expert and incorporates state-token masking to suppress visual and proprioceptive shortcuts during contact.
\end{itemize}

All methods use the same task demonstrations and corresponding evaluation configurations. Apart from their method-specific inputs and training strategies, the experimental conditions are kept consistent. For the two dual-expert methods in the main comparison, the learned automatic router described in Sec.~\ref{sec:contact_expert} is used for pipette tip attachment and bimanual pipette aspiration, whereas the transition to the contact expert is manually triggered for whiteboard wiping. The same routing protocol is used across the dual-expert variants for each task.

\noindent\textbf{Evaluation metrics.}
We report the overall task success rate together with the success rates of the non-contact and contact stages. The stage-wise metrics distinguish failures in approach and alignment from those occurring during force-sensitive interaction. A complete trial is successful only when both stages are completed.

For pipette tip attachment, success requires the tip to remain securely mounted after the micropipette retracts. For aspiration, success requires the robot to complete an effective aspiration action. For successful aspiration trials, we additionally measure the actual aspirated volume and compute its absolute error relative to the demonstration-derived target volume. For wiping, success requires the target marks to be removed while the eraser maintains effective contact with the board.

\noindent\textbf{Data and evaluation protocol.}
We collect 130 demonstrations for pipette tip attachment and 50 demonstrations each for bimanual pipette aspiration and whiteboard wiping. For the main comparison, every method is evaluated in 20 real-robot
trials per task. Before each trial, the relevant objects are randomly placed within a predefined workspace. The same set of test placements is used across methods to ensure a fair comparison.

\noindent\textbf{Implementation details.}
For pipette tip attachment, bimanual pipette aspiration, and
whiteboard wiping, the contact thresholds $\delta_{\mathrm{task}}$ are
$5.0$, $2.0$, and $4.0\,\mathrm{N}$, respectively, and the temporal
expansion parameters $(n,m)$ are $(10,25)$, $(20,20)$, and $(40,40)$
frames. We use $\sigma_{\max}=4.0$, $p_{\max}=0.5$,
$S_w=1{,}000$, and $S_d=29{,}000$. The contact experts and applicable
stage estimators are trained for 30,000 steps with a batch size of 8.

\begin{figure}[t]
    \centering
    \includegraphics[width=\columnwidth]
    {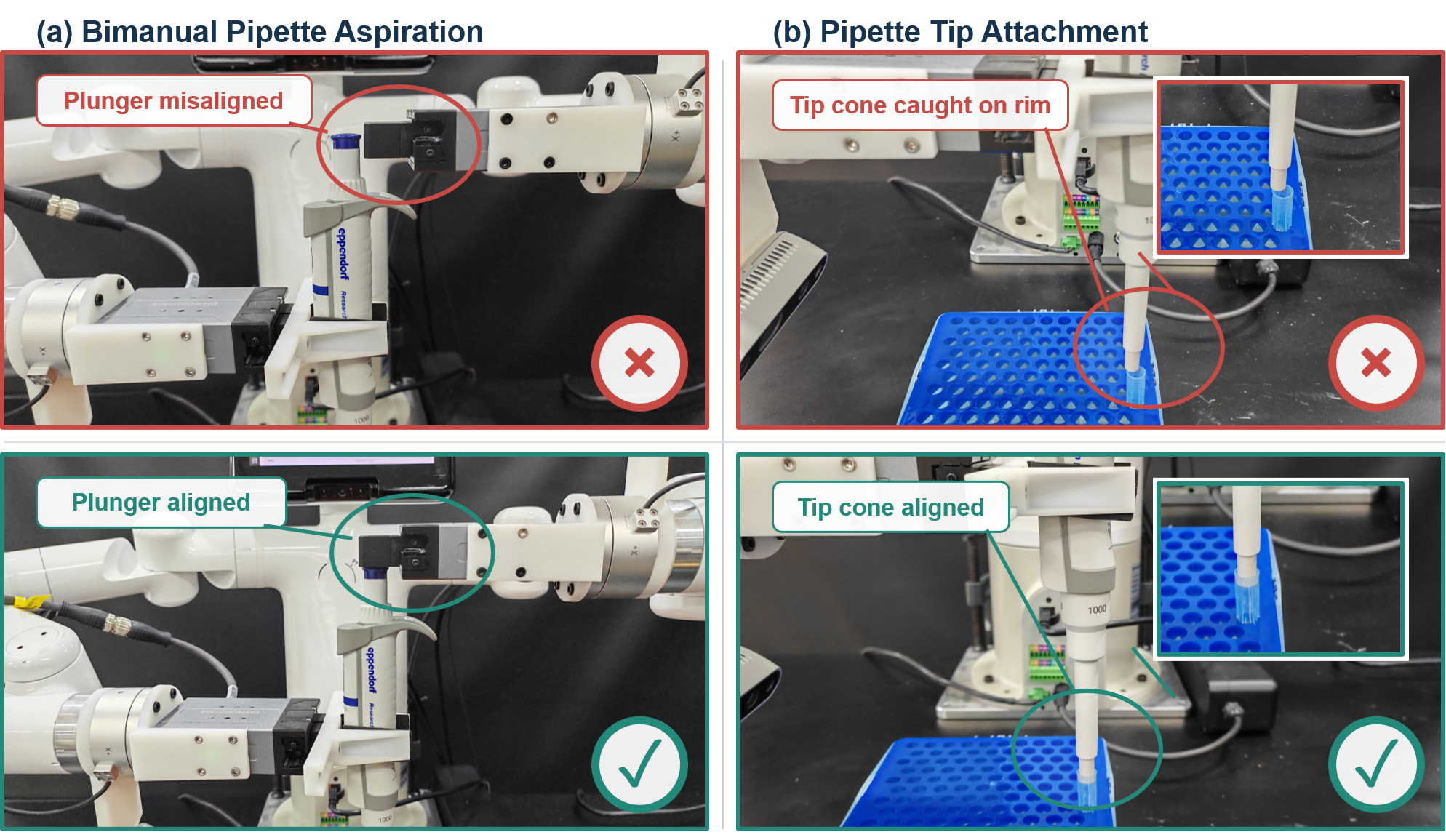}
    \caption{\textbf{Representative non-contact-stage outcomes.}
    The top and bottom rows show failed and successful visual alignment,
    respectively.
    (a) Alignment between the left end-effector and the pipette plunger
    button in bimanual aspiration.
    (b) Alignment between the pipette tip cone and the target tip
    opening in tip attachment.}
    \label{fig:noncontact_cases}
\end{figure}

\subsection{Main Results}
\label{sec:main_results}

\begin{table*}[t]
  \centering
  \caption{\textbf{Stage-wise and overall success rates across three
  contact-rich manipulation tasks.}}
  \label{tab:stagewise_task_success}
  \setlength{\tabcolsep}{3.0pt}
  \renewcommand{\arraystretch}{1.12}
  \resizebox{\textwidth}{!}{%
    \begin{tabular}{@{}lcccccccccc@{}}
      \toprule
      \multicolumn{1}{c}{\textbf{Method}} &
      \multicolumn{3}{c}{\textbf{Pipette tip attachment}} &
      \multicolumn{4}{c}{\textbf{Bimanual pipette aspiration}} &
      \multicolumn{3}{c}{\textbf{Whiteboard wiping}} \\
      \cmidrule(lr){2-4}
      \cmidrule(lr){5-8}
      \cmidrule(l){9-11}
      & \shortstack{Approach\\\& Align}
      & \shortstack{Press\\\& Retract}
      & Overall
      & \shortstack{Position\\\& Align}
      & \shortstack{Press\\\& Release}
      & Overall
      & \shortstack{Volume MAE\\($\mu\mathrm{L}$) $\downarrow$}
      & \shortstack{Approach\\\& Grasp}
      & \shortstack{Contact\\\& Wipe}
      & Overall \\
      \midrule
      $\pi_{0.5}$
        & \textbf{90\%} & 33.3\% & 30\%
        & 95\% & 5.3\% & 5\% & 120.0
        & \textbf{100\%} & 70\% & 70\% \\
      $\pi_{0.5}$ + Force Token
        & 40\% & 25\% & 10\%
        & \textbf{100\%} & 20\% & 20\% & 132.5
        & \textbf{100\%} & 45\% & 45\% \\
      $\pi_{0.5}$ + Force Token + Global FAC
        & 85\% & 29.4\% & 25\%
        & 35\% & 14.3\% & 5\% & 420.0
        & 50\% & 60\% & 30\% \\
      DEFT-VLA w/o Contact-Localized FAC
        & 80\% & 25\% & 20\%
        & \textbf{100\%} & 30\% & 30\% & 130.0
        & \textbf{100\%} & 55\% & 55\% \\
      \midrule
      \textbf{DEFT-VLA (Ours)}
        & 75\% & \textbf{93.3\%} & \textbf{70\%}
        & \textbf{100\%} & \textbf{50\%} & \textbf{50\%}
        & \textbf{104.0}
        & \textbf{100\%} & \textbf{90\%} & \textbf{90\%} \\
      \bottomrule
    \end{tabular}%
  }
  
    \par\vspace{4pt}
    \noindent
    \begin{minipage}{\textwidth}
     \raggedright
     \normalfont\footnotesize
    Non-contact and overall rates use all 20 trials; contact-stage rates
    are conditioned on successful completion of the preceding non-contact
    stage. Volume MAE is the mean absolute deviation from the
    $1000\,\mu\mathrm{L}$ reference over successful aspiration trials;
    lower is better. Boldface indicates the best result in each column,
    including ties.
  \end{minipage}
\end{table*}

Table~\ref{tab:stagewise_task_success} reports the stage-wise and overall
performance of all methods. DEFT-VLA achieves the highest overall
success rate across all three tasks, reaching $70\%$ on pipette tip
attachment, $50\%$ on bimanual pipette aspiration, and $90\%$ on
whiteboard wiping. These results exceed the best alternative in each
task by $40$, $20$, and $20$ percentage points, respectively. The
improvements are concentrated primarily in the force-sensitive contact
stage. Conditioned on successful completion of the preceding
vision-dominated stage, DEFT-VLA achieves contact-stage success rates of
$14/15$ ($93.3\%$), $10/20$ ($50\%$), and $18/20$ ($90\%$).

The task-adapted $\pi_{0.5}$ retains strong performance during the
vision-dominated stage, achieving non-contact success rates of $90\%$,
$95\%$, and $100\%$, but performs considerably worse during contact.
Directly adding a force token does not provide consistent gains: it
improves aspiration performance, but reduces the overall success rate
on tip attachment and wiping relative to $\pi_{0.5}$. Applying the
Force-Attending Curriculum globally to a single expert is similarly
inconsistent. In particular, its non-contact success rate drops to
$35\%$ on aspiration and $50\%$ on wiping. We observe $13$ positioning or alignment failures in aspiration and
$10$ visual grasping failures in wiping. Representative non-contact
outcomes are shown in Fig.~\ref{fig:noncontact_cases}: misalignment with
the pipette plunger prevents effective bimanual aspiration, while an
offset between the pipette tip cone and the target opening causes the
cone to become caught on the rim. These cases indicate that applying
force-attending training over complete task trajectories can interfere
with preparation behaviors that remain strongly dependent on vision.

Comparing the two dual-expert variants further shows that the benefit
of the complete training strategy arises mainly during contact.
DEFT-VLA without Contact-Localized FAC achieves non-contact success
rates of $80\%$, $100\%$, and $100\%$, but its contact-stage success
rates are only $25\%$, $30\%$, and $55\%$. With the contact-localized
force-attending curriculum and state-token masking, the corresponding
non-contact success rates remain comparable at $75\%$, $100\%$, and
$100\%$, while contact-stage success increases to $93.3\%$, $50\%$,
and $90\%$. These results show that DEFT-VLA substantially improves
force-sensitive interaction while largely preserving the visuomotor
competence of the task-adapted base expert.

The observed failure modes further illustrate the differences between
the methods. In tip attachment, contact failures of the baseline
methods frequently involve insufficient insertion force, tip
detachment, or repeated insertion and retraction. In contrast,
DEFT-VLA successfully completes $14$ of the $15$ trials that reach the
contact stage. In whiteboard wiping, DEFT-VLA succeeds in $18/20$
trials, with only two
failures caused by excessive contact force. In aspiration, DEFT-VLA
completes $10$ effective aspiration trials, compared with $6$ for the
best alternative, and achieves the lowest mean absolute volume error
of $104.0\,\mu\mathrm{L}$ over successful trials. The remaining
failures mainly involve repeated pressing or mistimed expert
switching.

\subsection{Model Generalization}
\label{sec:height_generalization}

\begin{figure}[t]
  \centering
  \includegraphics[width=0.82\columnwidth]
  {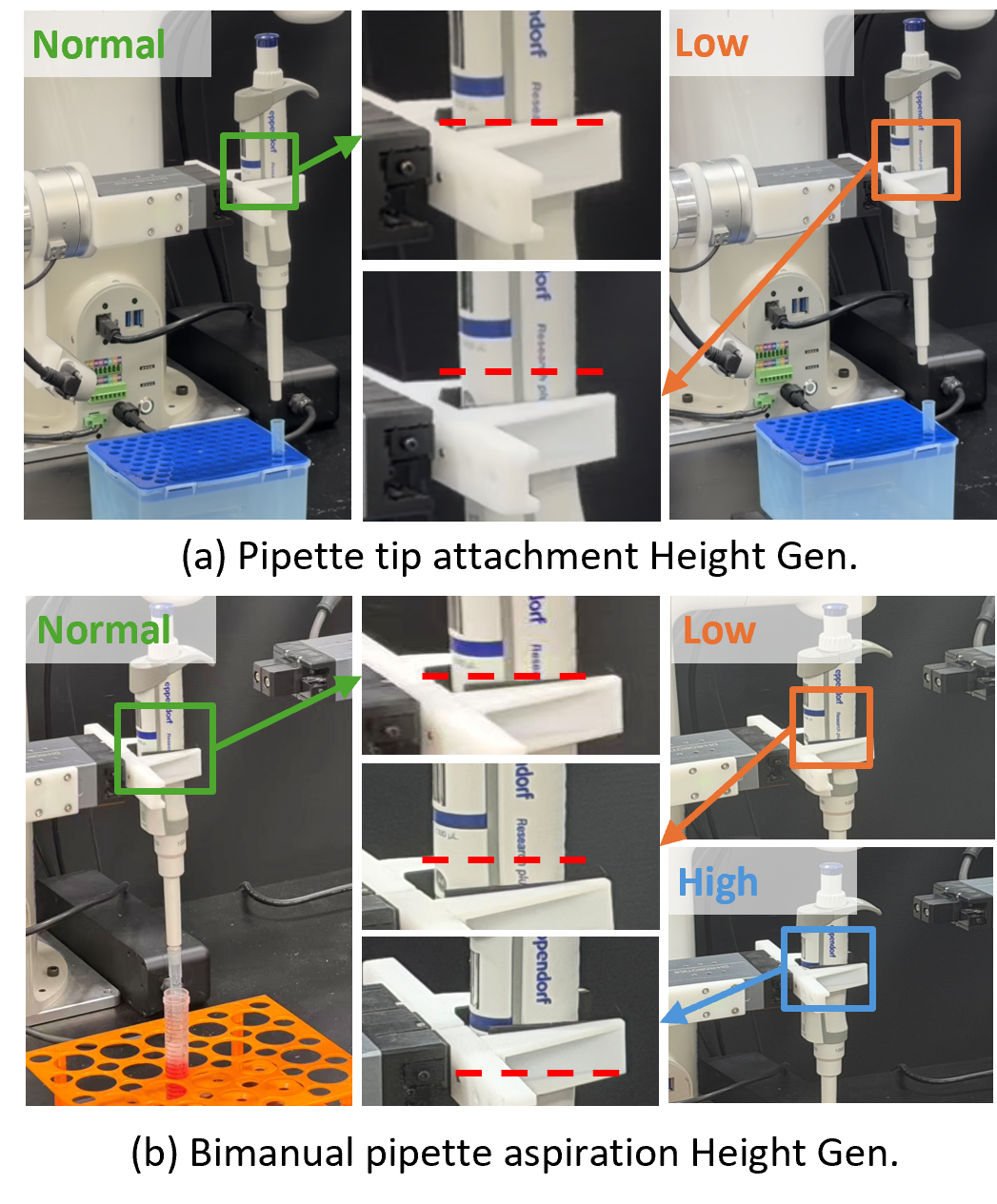}

  \setlength{\abovecaptionskip}{2pt}
  \caption{\textbf{Height generalization protocol.}
  Height labels refer to the gripper's grasp position along the pipette.
  (a) Tip attachment at the \emph{Normal} and held-out \emph{Low}
  positions. (b) Bimanual aspiration at the \emph{Normal} and held-out
  \emph{Low} and \emph{High} positions. Red dashed lines indicate the
  \emph{Normal} reference in the magnified views.}
  \label{fig:height_generalization}
\end{figure}

\begin{table}[t]
  \centering
  \caption{\textbf{Success rates under height generalization.}}
  \label{tab:height_generalization}
  \setlength{\tabcolsep}{2.2pt}
  \renewcommand{\arraystretch}{1.08}

  \resizebox{0.98\columnwidth}{!}{%
    \begin{tabular}{@{}lcc@{}}
      \toprule
      \multicolumn{1}{c}{\textbf{Method}} &
      \textbf{Pipette tip attachment} &
      \textbf{Bimanual pipette aspiration} \\
      \midrule
      $\pi_{0.5}$ & 0\% & 0\% \\
      $\pi_{0.5}$ + Force Token & 10\% & 10\% \\
      $\pi_{0.5}$ + Force Token + Global FAC & 0\% & 0\% \\
      DEFT-VLA w/o Contact-Localized FAC & 0\% & 40\% \\
      \midrule
      \textbf{DEFT-VLA (Ours)}
      & \textbf{50\%} & \textbf{50\%} \\
      \bottomrule
    \end{tabular}%
  }

  \vspace{4pt}
  \parbox{\columnwidth}{\normalfont\footnotesize
  Results are evaluated at held-out gripper grasp positions. For
  bimanual aspiration, the reported result combines the \emph{High}
  and \emph{Low} conditions. Boldface indicates the best result in
  each column, including ties.}
\end{table}



To further evaluate generalization beyond the demonstrated physical
configurations, we vary the initial gripper--pipette grasp height in
the tip-attachment and bimanual-aspiration tasks. All evaluations are
conducted at grasp heights absent from the demonstrations. As
illustrated in Fig.~\ref{fig:height_generalization}, tip attachment is
evaluated using a lower grasp position, whereas aspiration includes
both higher and lower grasp positions. The Normal grasp height in the
figure is shown only as the training-distribution reference and is not
included in the results. For each method, we conduct 10 trials per
task: all 10 tip-attachment trials use the Low condition, while the
aspiration trials comprise five High and five Low trials, whose
combined success rate is reported in
Table~\ref{tab:height_generalization}. These variations shift the
pipette tip and plunger button relative to the end-effectors, requiring
the policy to adjust its approach, contact position, and pressing
distance rather than simply reproduce the demonstrated trajectories.

As shown in Table~\ref{tab:height_generalization}, DEFT-VLA succeeds in
5 of 10 trials ($50\%$) on both tasks. For tip attachment, all baselines
fail completely except $\pi_{0.5}$ + Force Token, which succeeds in one
trial ($10\%$). For bimanual aspiration, the single-expert baselines
achieve at most $10\%$, while DEFT-VLA w/o Contact-Localized FAC
reaches $40\%$. These results suggest that retaining the task-adapted
base expert for visual positioning and alignment, while specializing
the contact expert through contact-localized force-attending training,
improves adaptation to the positional and motion displacements caused
by unseen grasp heights.

\FloatBarrier

\subsection{State-Token Masking Ablation}
\label{sec:state_masking_ablation}

\noindent\begin{minipage}{\columnwidth}
    \centering
    \captionof{table}{\textbf{Ablation of state-token masking on pipette tip
    attachment.}}
    \label{tab:state_masking_ablation}
    \resizebox{\columnwidth}{!}{
    \begin{tabular}{lccc}
        \toprule
        \textbf{Method}
        & \textbf{Visual Curriculum}
        & \textbf{State Masking}
        & \textbf{Success Rate} \\
        \midrule
        DEFT-VLA w/o Contact-Localized FAC
        & -- & -- & 20\% \\
        DEFT-VLA w/o State Masking
        & $\checkmark$ & -- & 15\% \\
        \midrule
        \textbf{DEFT-VLA (Ours)}
        & $\checkmark$ & $\checkmark$ & \textbf{70\%} \\
        \bottomrule
    \end{tabular}
    }
\end{minipage}

\par\vspace{0.6em}

To examine the effect of state-token masking, we compare three
dual-expert variants on pipette tip attachment, evaluating each
variant over 20 trials under the same protocol. As shown in
Table~\ref{tab:state_masking_ablation}, DEFT-VLA without the entire
Contact-Localized FAC achieves a success rate of $20\%$, while
retaining the visual curriculum but removing state-token masking
results in $15\%$. The difference corresponds to only one successful
trial, indicating that suppressing visual information alone provides
no clear improvement. In contrast, the complete DEFT-VLA achieves
$70\%$, outperforming the direct variant without state-token masking
by $55$ percentage points. This result suggests that, when visual cues
are suppressed, robot states may remain as alternative shortcuts for
contact-action prediction. Masking a subset of the state tokens helps
reduce these shortcuts and encourages the contact expert to make more
effective use of force feedback.

\FloatBarrier

\section{Conclusion}

This work presented DEFT-VLA, a stage-localized approach to force-aware post-training of VLA policies. Rather than globally adapting a single policy to force feedback, DEFT-VLA retains a task-adapted base expert for vision-dominated preparation and specializes a separate force-aware expert using contact-stage demonstrations. Its contact-localized force-attending curriculum progressively relaxes visual suppression and state-token masking, encouraging effective force utilization without modifying the base expert. Across three real-robot tasks, DEFT-VLA improves overall success by 20--40 percentage points over the strongest baseline on each task, with gains concentrated in force-sensitive contact stages while maintaining strong non-contact performance. It further achieves a 50\% success rate under held-out grasp heights on both pipette tasks, while the ablation supports the importance of state-token masking in reducing proprioceptive shortcuts. These results demonstrate the value of localizing sensory adaptation to task stages where the corresponding feedback is most informative. Current limitations include evaluation on three task families and the
use of manual expert switching for whiteboard wiping; future work will
study more general automatic routing and broader cross-task transfer.









\bibliographystyle{IEEEtran}
\bibliography{references}

@IEEEtranBSTCTL{IEEEexample:BSTcontrol,
  CTLuse_forced_etal       = "yes",
  CTLmax_names_forced_etal = "6",
  CTLnames_show_etal       = "1",
  CTLdash_repeated_names   = "no"
}

@inproceedings{brohan2023rt2,
  author    = {Brianna Zitkovich and Tianhe Yu and Sichun Xu and Peng Xu
               and Ted Xiao and Fei Xia and Jialin Wu and Paul Wohlhart
               and Stefan Welker and Ayzaan Wahid and Quan Vuong
               and Vincent Vanhoucke and Huong Tran and Radu Soricut
               and Anikait Singh and Jaspiar Singh and Pierre Sermanet
               and Pannag R. Sanketi and Grecia Salazar and Michael S. Ryoo
               and Krista Reymann and Kanishka Rao and Karl Pertsch
               and Igor Mordatch and Henryk Michalewski and Yao Lu
               and Sergey Levine and Lisa Lee and Tsang-Wei Edward Lee
               and Isabel Leal and Yuheng Kuang and Dmitry Kalashnikov
               and Ryan Julian and Nikhil J. Joshi and Alex Irpan
               and Brian Ichter and Jasmine Hsu and Alexander Herzog
               and Karol Hausman and Keerthana Gopalakrishnan and Chuyuan Fu
               and Pete Florence and Chelsea Finn and Kumar Avinava Dubey
               and Danny Driess and Tianli Ding and Krzysztof Marcin Choromanski
               and Xi Chen and Yevgen Chebotar and Justice Carbajal
               and Noah Brown and Anthony Brohan and Montserrat Gonzalez Arenas
               and Kehang Han},
  title     = {{RT-2}: Vision-Language-Action Models Transfer Web Knowledge to Robotic Control},
  booktitle = {Proc. 7th Conf. Robot Learn. (CoRL)},
  volume    = {229},
  pages     = {2165--2183},
  year      = {2023},
  url       = {https://proceedings.mlr.press/v229/zitkovich23a.html}
}

@inproceedings{kim2025openvla,
  author    = {Moo Jin Kim and Karl Pertsch and Siddharth Karamcheti
               and Ted Xiao and Ashwin Balakrishna and Suraj Nair
               and Rafael Rafailov and Ethan P. Foster and Pannag R. Sanketi
               and Quan Vuong and Thomas Kollar and Benjamin Burchfiel
               and Russ Tedrake and Dorsa Sadigh and Sergey Levine
               and Percy Liang and Chelsea Finn},
  title     = {{OpenVLA}: An Open-Source Vision-Language-Action Model},
  booktitle = {Proc. 8th Conf. Robot Learn. (CoRL)},
  volume    = {270},
  pages     = {2679--2713},
  year      = {2025},
  url       = {https://proceedings.mlr.press/v270/kim25c.html}
}

@inproceedings{black2024pi0,
  author    = {Kevin Black and Noah Brown and Danny Driess and Adnan Esmail
               and Michael Robert Equi and Chelsea Finn and Niccolo Fusai
               and Lachy Groom and Karol Hausman and Brian Ichter
               and Szymon Jakubczak and Tim Jones and Liyiming Ke
               and Sergey Levine and Adrian Li-Bell and Mohith Mothukuri
               and Suraj Nair and Karl Pertsch and Lucy Xiaoyang Shi
               and Laura Smith and James Tanner and Quan Vuong
               and Anna Walling and Haohuan Wang and Ury Zhilinsky},
  title     = {{$\pi_0$: A Vision-Language-Action Flow Model for General Robot Control}},
  booktitle = {Proc. Robot.: Sci. Syst. (RSS)},
  year      = {2025},
  doi       = {10.15607/RSS.2025.XXI.010},
  note      = {doi: 10.15607/RSS.2025.XXI.010}
}

@inproceedings{black2025pi05,
  author    = {Kevin Black and Noah Brown and James Darpinian and Karan Dhabalia
               and Danny Driess and Adnan Esmail and Michael Robert Equi
               and Chelsea Finn and Niccolo Fusai and Manuel Y. Galliker
               and Dibya Ghosh and Lachy Groom and Karol Hausman and Brian Ichter
               and Szymon Jakubczak and Tim Jones and Liyiming Ke and Devin LeBlanc
               and Sergey Levine and Adrian Li-Bell and Mohith Mothukuri
               and Suraj Nair and Karl Pertsch and Allen Z. Ren
               and Lucy Xiaoyang Shi and Laura Smith and Jost Tobias Springenberg
               and Kyle Stachowicz and James Tanner and Quan Vuong and Homer Walke
               and Anna Walling and Haohuan Wang and Lili Yu and Ury Zhilinsky},
  title     = {{$\pi_{0.5}$: a Vision-Language-Action Model with Open-World Generalization}},
  booktitle = {Proc. 9th Conf. Robot Learn. (CoRL)},
  volume    = {305},
  pages     = {17--40},
  year      = {2025},
  url       = {https://proceedings.mlr.press/v305/black25a.html}
}

@inproceedings{liu2025forcemimic,
  author    = {Wenhai Liu and Junbo Wang and Yiming Wang and Weiming Wang and Cewu Lu},
  title     = {{ForceMimic}: Force-Centric Imitation Learning with Force-Motion Capture System for Contact-Rich Manipulation},
  booktitle = {Proc. IEEE Int. Conf. Robot. Autom. (ICRA)},
  pages     = {1105--1112},
  year      = {2025},
  doi       = {10.1109/ICRA55743.2025.11128061},
  note      = {doi: 10.1109/ICRA55743.2025.11128061}
}

@article{he2025foar,
  author  = {Zihao He and Hongjie Fang and Jingjing Chen and Hao-Shu Fang and Cewu Lu},
  title   = {{FoAR}: Force-Aware Reactive Policy for Contact-Rich Robotic Manipulation},
  journal = {IEEE Robot. Autom. Lett.},
  volume  = {10},
  number  = {6},
  pages   = {5625--5632},
  month   = jun,
  year    = {2025},
  doi     = {10.1109/LRA.2025.3560871},
  note    = {doi: 10.1109/LRA.2025.3560871}
}

@inproceedings{xue2025reactive,
  author    = {Han Xue and Jieji Ren and Wendi Chen and Gu Zhang
               and Yuan Fang and Guoying Gu and Huazhe Xu and Cewu Lu},
  title     = {Reactive Diffusion Policy: Slow-Fast Visual-Tactile Policy Learning for Contact-Rich Manipulation},
  booktitle = {Proc. Robot.: Sci. Syst. (RSS)},
  year      = {2025},
  doi       = {10.15607/RSS.2025.XXI.052},
  note      = {doi: 10.15607/RSS.2025.XXI.052}
}

@inproceedings{yu2025forcevla,
  author    = {Jiawen Yu and Hairuo Liu and Qiaojun Yu and Jieji Ren
               and Ce Hao and Haitong Ding and Guangyu Huang and Guofan Huang
               and Yan Song and Panpan Cai and Wenqiang Zhang and Cewu Lu},
  title     = {{ForceVLA}: Enhancing {VLA} Models with a Force-Aware {MoE} for Contact-Rich Manipulation},
  booktitle = {Adv. Neural Inf. Process. Syst.},
  volume    = {38},
  pages     = {103609--103639},
  year      = {2025},
  doi       = {10.52202/085713-3124},
  note      = {doi: 10.52202/085713-3124}
}

@inproceedings{liu2025factr,
  author    = {Jason Jingzhou Liu and Yulong Li and Kenneth Shaw and Tony Tao
               and Ruslan Salakhutdinov and Deepak Pathak},
  title     = {{FACTR}: Force-Attending Curriculum Training for Contact-Rich Policy Learning},
  booktitle = {Proc. Robot.: Sci. Syst. (RSS)},
  year      = {2025},
  doi       = {10.15607/RSS.2025.XXI.079},
  note      = {doi: 10.15607/RSS.2025.XXI.079}
}

@misc{zhang2026craft,
  author = {Yike Zhang and Yaonan Wang and Xinxin Sun and Kaizhen Huang
            and Zhiyuan Xu and Junjie Ji and Zhengping Che and Jian Tang
            and Jingtao Sun},
  title  = {{CRAFT}: Adapting {VLA} Models to Contact-Rich Manipulation via Force-Aware Curriculum Fine-Tuning},
  year   = {2026},
  note   = {arXiv:2602.12532}
}

@inproceedings{hou2024adaptivecompliance,
  author    = {Yifan Hou and Zeyi Liu and Cheng Chi and Eric Cousineau
               and Naveen Kuppuswamy and Siyuan Feng and Benjamin Burchfiel
               and Shuran Song},
  title     = {Adaptive Compliance Policy: Learning Approximate Compliance for Diffusion Guided Control},
  booktitle = {Proc. IEEE Int. Conf. Robot. Autom. (ICRA)},
  pages     = {4829--4836},
  year      = {2025},
  doi       = {10.1109/ICRA55743.2025.11128452},
  note      = {doi: 10.1109/ICRA55743.2025.11128452}
}

@inproceedings{tacdiffusion,
  author    = {Yansong Wu and Zongxie Chen and Fan Wu and Lingyun Chen
               and Liding Zhang and Zhenshan Bing and Abdalla Swikir
               and Sami Haddadin and Alois Knoll},
  title     = {{TacDiffusion}: Force-Domain Diffusion Policy for Precise Tactile Manipulation},
  booktitle = {Proc. IEEE Int. Conf. Robot. Autom. (ICRA)},
  pages     = {11831--11837},
  year      = {2025},
  doi       = {10.1109/ICRA55743.2025.11127334},
  note      = {doi: 10.1109/ICRA55743.2025.11127334}
}

@article{visualforcetactile,
  author  = {Piaopiao Jin and Bidan Huang and Wang Wei Lee and Tiefeng Li and Wei Yang},
  title   = {Visual-Force-Tactile Fusion for Gentle Intricate Insertion Tasks},
  journal = {IEEE Robot. Autom. Lett.},
  volume  = {9},
  number  = {5},
  pages   = {4830--4837},
  month   = may,
  year    = {2024},
  doi     = {10.1109/LRA.2024.3379803},
  note    = {doi: 10.1109/LRA.2024.3379803}
}

@article{multimodalforcematched,
  author  = {Trevor Ablett and Oliver Limoyo and Adam Sigal and Affan Jilani
             and Jonathan Kelly and Kaleem Siddiqi and Francois Hogan
             and Gregory Dudek},
  title   = {Multimodal and Force-Matched Imitation Learning With a See-Through Visuotactile Sensor},
  journal = {IEEE Trans. Robot.},
  volume  = {41},
  pages   = {946--959},
  year    = {2025},
  doi     = {10.1109/TRO.2024.3521864},
  note    = {doi: 10.1109/TRO.2024.3521864}
}

@inproceedings{zhang2025tavla,
  author    = {Zongzheng Zhang and Haobo Xu and Zhuo Yang and Chenghao Yue
               and Zehao Lin and Huan-{ang} Gao and Ziwei Wang and Hao Zhao},
  title     = {Elucidating the Design Space of Torque-Aware Vision-Language-Action Models},
  booktitle = {Proc. 9th Conf. Robot Learn. (CoRL)},
  volume    = {305},
  pages     = {4019--4037},
  year      = {2025},
  url       = {https://proceedings.mlr.press/v305/zhang25k.html}
}

@inproceedings{li2026forcevla2,
  author    = {Yang Li and {{Zhaxizhuoma}} and Hongru Jiang
               and Junjie Xia and Hongquan Zhang and Jinda Du
               and Yunsong Zhou and Jia Zeng and Ce Hao and Jieji Ren
               and Qiaojun Yu and Cewu Lu and Yu Qiao and Jiangmiao Pang},
  title     = {{ForceVLA2}: Unleashing Hybrid Force-Position Control
               with Force Awareness for Contact-Rich Manipulation},
  booktitle = {Proc. IEEE/CVF Conf. Comput. Vis. Pattern Recognit. (CVPR)},
  pages     = {8911--8920},
  month     = jun,
  year      = {2026}
}

@misc{huang2025tactilevla,
  author = {Jialei Huang and Shuo Wang and Fanqi Lin and Yihang Hu
            and Chuan Wen and Yang Gao},
  title  = {{Tactile-VLA}: Unlocking Vision-Language-Action Model's Physical Knowledge for Tactile Generalization},
  year   = {2025},
  note   = {arXiv:2507.09160}
}

@misc{zhang2026tacvla,
  author = {Kaidi Zhang and Heng Zhang and Zhengtong Xu and Zhiyuan Zhang
            and Md Rakibul Islam Prince and Xiang Li and Xiaojing Han
            and Yuhao Zhou and Arash Ajoudani and Yu She},
  title  = {{TacVLA}: Contact-Aware Tactile Fusion for Robust Vision-Language-Action Manipulation},
  year   = {2026},
  note   = {arXiv:2603.12665}
}

@misc{li2026favla,
  author = {Yao Li and Peiyuan Tang and Wuyang Zhang and Chengyang Zhu
            and Yifan Duan and Weikai Shi and Xiaodong Zhang and Zijiang Yang
            and Jianmin Ji and Yanyong Zhang},
  title  = {{FAVLA}: A Force-Adaptive Fast-Slow {VLA} Model for Contact-Rich Robotic Manipulation},
  year   = {2026},
  note   = {arXiv:2602.23648}
}

@misc{wang2026lift,
  author = {Yi Wang and Wendi Chen and Zimo Wen and Han Xue and Xueqi Li
            and Wenye Yu and Zhijie Chen and Hao Yang and Jun Lv
            and Chuan Wen and Cewu Lu},
  title  = {Never Too Late for Force: Accelerating {VLA} Post-Training with Reactive Force Injection},
  year   = {2026},
  note   = {arXiv:2607.14236}
}

@misc{forceaware_dagger,
  author = {Yiou Huang and Ning Ma and Weichu Zhao and Zinuo Liu
            and Jun Sun and Qiufeng Wang and Yaran Chen},
  title  = {Force-Aware Residual {DAgger} via Trajectory Editing for Precision Insertion with Impedance Control},
  year   = {2026},
  note   = {arXiv:2603.04038}
}

@misc{zheng2026torlvla,
  author = {Huaihang Zheng and Yi Yang and Kai Ma and Shenglin Xu
            and Tian Xie and Guozheng Li and Xiangyu Wang and Yiren Ma
            and Si Liu and Yinian Mao and Baoxu Liu},
  title  = {{TORL-VLA}: Tactile Guided Online Reinforcement Learning for Contact-Rich Manipulation},
  year   = {2026},
  note   = {arXiv:2606.09337}
}

@misc{oh2026factr2,
  author = {Steven Oh and Jason Jingzhou Liu and Tony Tao and Philip Han
            and Kenneth Shaw and Satoshi Funabashi and Ruslan Salakhutdinov
            and Deepak Pathak},
  title  = {{FACTR 2}: Learning External Force Sensing for Commodity Robot Arms Improves Policy Learning},
  year   = {2026},
  note   = {arXiv:2606.12406}
}

@inproceedings{brohan2023rt1,
  author    = {Anthony Brohan and Noah Brown and Justice Carbajal and Yevgen Chebotar
               and Joseph Dabis and Chelsea Finn and Keerthana Gopalakrishnan
               and Karol Hausman and Alexander Herzog and Jasmine Hsu and Julian Ibarz
               and Brian Ichter and Alex Irpan and Tomas Jackson and Sally Jesmonth
               and Nikhil Joshi and Ryan Julian and Dmitry Kalashnikov and Yuheng Kuang
               and Isabel Leal and Kuang-Huei Lee and Sergey Levine and Yao Lu
               and Utsav Malla and Deeksha Manjunath and Igor Mordatch and Ofir Nachum
               and Carolina Parada and Jodilyn Peralta and Emily Perez and Karl Pertsch
               and Jornell Quiambao and Kanishka Rao and Michael S. Ryoo and Grecia Salazar
               and Pannag R. Sanketi and Kevin Sayed and Jaspiar Singh and Sumedh Sontakke
               and Austin Stone and Clayton Tan and Huong Tran and Vincent Vanhoucke
               and Steve Vega and Quan H. Vuong and Fei Xia and Ted Xiao and Peng Xu
               and Sichun Xu and Tianhe Yu and Brianna Zitkovich},
  title     = {{RT-1}: Robotics Transformer for Real-World Control at Scale},
  booktitle = {Proc. Robot.: Sci. Syst. (RSS)},
  month     = jul,
  year      = {2023},
  doi       = {10.15607/RSS.2023.XIX.025},
  note      = {doi: 10.15607/RSS.2023.XIX.025}
}

@inproceedings{openx2024rtx,
  author    = {{Open X-Embodiment Collaboration}},
  title     = {Open {X}-Embodiment: Robotic Learning Datasets and {RT-X} Models},
  booktitle = {Proc. IEEE Int. Conf. Robot. Autom. (ICRA)},
  pages     = {6892--6903},
  month     = may,
  year      = {2024},
  doi       = {10.1109/ICRA57147.2024.10611477},
  note      = {doi: 10.1109/ICRA57147.2024.10611477}
}

@inproceedings{ghosh2024octo,
  author    = {Dibya Ghosh and Homer Rich Walke and Karl Pertsch and Kevin Black
               and Oier Mees and Sudeep Dasari and Joey Hejna and Tobias Kreiman
               and Charles Xu and Jianlan Luo and You Liang Tan and Lawrence Yunliang Chen
               and Quan Vuong and Ted Xiao and Pannag R. Sanketi and Dorsa Sadigh
               and Chelsea Finn and Sergey Levine},
  title     = {{Octo}: An Open-Source Generalist Robot Policy},
  booktitle = {Proc. Robot.: Sci. Syst. (RSS)},
  month     = jul,
  year      = {2024},
  doi       = {10.15607/RSS.2024.XX.090},
  note      = {doi: 10.15607/RSS.2024.XX.090}
}

@inproceedings{zhao2023act,
  author    = {Tony Z. Zhao and Vikash Kumar and Sergey Levine and Chelsea Finn},
  title     = {Learning Fine-Grained Bimanual Manipulation with Low-Cost Hardware},
  booktitle = {Proc. Robot.: Sci. Syst. (RSS)},
  month     = jul,
  year      = {2023},
  doi       = {10.15607/RSS.2023.XIX.016},
  note      = {doi: 10.15607/RSS.2023.XIX.016}
}

@inproceedings{chi2023diffusion,
  author    = {Cheng Chi and Siyuan Feng and Yilun Du and Zhenjia Xu
               and Eric Cousineau and Benjamin C. M. Burchfiel and Shuran Song},
  title     = {Diffusion Policy: Visuomotor Policy Learning via Action Diffusion},
  booktitle = {Proc. Robot.: Sci. Syst. (RSS)},
  month     = jul,
  year      = {2023},
  doi       = {10.15607/RSS.2023.XIX.026},
  note      = {doi: 10.15607/RSS.2023.XIX.026}
}

@inproceedings{fang2026forcepolicy,
  author    = {Hongjie Fang and Shirun Tang and Mingyu Mei and Haoxiang Qin
               and Zihao He and Jingjing Chen and Ying Feng and Chenxi Wang
               and Wanxi Liu and Zaixing He and Cewu Lu and Shiquan Wang},
  title     = {Force Policy: Learning Hybrid Force-Position Control Policy under Interaction Frame for Contact-Rich Manipulation},
  booktitle = {Proc. Robot.: Sci. Syst. (RSS)},
  month     = jul,
  year      = {2026},
  doi       = {10.15607/RSS.2026.XXII.128},
  note      = {doi: 10.15607/RSS.2026.XXII.128}
}

@inproceedings{dehaan2019causalconfusion,
  author    = {Pim de Haan and Dinesh Jayaraman and Sergey Levine},
  title     = {Causal Confusion in Imitation Learning},
  booktitle = {Advances in Neural Information Processing Systems},
  volume    = {32},
  year      = {2019}
}

@misc{driess2025knowledgeinsulation,
  author = {Danny Driess and Jost Tobias Springenberg and Brian Ichter
            and Lili Yu and Adrian Li-Bell and Karl Pertsch
            and Allen Z. Ren and Homer Walke and Quan Vuong
            and Lucy Xiaoyang Shi and Sergey Levine},
  title  = {Knowledge Insulating Vision-Language-Action Models:
            Train Fast, Run Fast, Generalize Better},
  year   = {2025},
  note   = {arXiv:2505.23705}
}

@misc{guo2026priorvla,
  author = {Xinyu Guo and Bin Xie and Wei Chai and Xianchi Deng
            and Tiancai Wang and Zhengxing Wu and Xingyu Chen},
  title  = {{PriorVLA}: Prior-Preserving Adaptation for
            Vision-Language-Action Models},
  year   = {2026},
  note   = {arXiv:2605.10925}
}

@misc{li2026fmvla,
  author = {Ruicheng Li and Qixiu Li and Ruichun Ma and Yu Deng
            and Lin Luo and Zhiying Du and Jianfeng Xiang
            and Huizhi Liang and Ruicheng Wang and Jiaolong Yang
            and Baining Guo},
  title  = {{FM-VLA}: Force-Based Memory for Vision-Language-Action
            Models in Contact-Rich Manipulation},
  year   = {2026},
  note   = {arXiv:2607.18231}
}

@misc{zhang2026fwbcvla,
  author = {Yutian Zhang and Siyuan Ma and Liwen Yang and Yang Li
            and Ce Hao and Haozhen Chi and Dong We and Qiaojun Yu
            and Dibo Hou},
  title  = {{FWBC-VLA}: Force-Aware Whole-Body Compensation for
            Contact-Rich Loco-Manipulation},
  year   = {2026},
  note   = {arXiv:2609.03889}
}

\end{document}